\pdfoutput=1
\documentclass[11pt]{article}

\usepackage{EMNLP2023}

\usepackage{times}
\usepackage{latexsym}

\usepackage[T1]{fontenc}
\usepackage[utf8]{inputenc}
\usepackage{microtype}

\usepackage{inconsolata}
\usepackage{graphicx}
\usepackage{amsmath}
\usepackage{amssymb}
\usepackage{algorithm}
\usepackage{algpseudocode}
\usepackage{comment}
\usepackage{graphicx}
\usepackage{multirow}
\usepackage{multicol}
\usepackage{array}
\usepackage{color, colortbl}
\usepackage{xcolor}
\usepackage{enumitem}

\definecolor{bg}{rgb}{0.95, 0.95, 0.95}
\definecolor{mbg}{rgb}{0.75, 0.75, 0.75}
\definecolor{mmbg}{rgb}{0.65, 0.65, 0.65}

\usepackage{hyperref}

\def\hyphenateAndTtWholeString #1{\xHyphenate#1$\wholeString\unskip}

\def\xHyphenate#1#2\wholeString {\if#1$%
    \else\transform{#1}%
    \takeTheRest#2\ofTheString\fi}

\def\takeTheRest#1\ofTheString\fi
{\fi \xHyphenate#1\wholeString}

\def\transform#1{\url{#1}\hskip 0pt plus 1pt}

\def\urlx #1{\href{#1}{\hyphenateAndTtWholeString{#1}}}

\title{CulturaX: A Cleaned, Enormous, and Multilingual Dataset for Large Language Models in 167 Languages}

\author{{\bf Thuat Nguyen$^1$}, Chien Van Nguyen$^{1}$, Viet Dac Lai$^{1}$, Hieu Man$^{1}$, Nghia Trung Ngo$^{1}$ \\
        {\bf Franck Dernoncourt$^2$, Ryan A. Rossi$^2$, Thien Huu Nguyen$^1$ } \\
        $^1$Dept. of Computer Science, University of Oregon, OR, USA\\
        $^2$Adobe Research, USA\\
       \texttt{nguyenhuuthuat09@gmail.com}
       \\\texttt{\{chienn,vietl@cs,hieum,nghian,thien@cs\}@uoregon.edu}\\
       \texttt{\{franck.dernoncourt,ryrossi\}@adobe.com}
      }

\begin{document}
\maketitle
\begin{abstract}

The driving factors behind the development of large language models (LLMs) with impressive learning capabilities are their colossal model sizes and extensive training datasets. Along with the progress in natural language processing, LLMs have been frequently made accessible to the public to foster deeper investigation and applications. However, when it comes to training datasets for these LLMs, especially the recent state-of-the-art models, they are often not fully disclosed. Creating training data for high-performing LLMs involves extensive cleaning and deduplication to ensure the necessary level of quality. The lack of transparency for training data has thus hampered research on attributing and addressing hallucination and bias issues in LLMs, hindering replication efforts and further advancements in the community. These challenges become even more pronounced in multilingual learning scenarios, where the available multilingual text datasets are often inadequately collected and cleaned. Consequently, there is a lack of open-source and readily usable dataset to effectively train LLMs in multiple languages. To overcome this issue, we present CulturaX, a substantial multilingual dataset with 6.3 trillion tokens in 167 languages, tailored for LLM development. Our dataset undergoes meticulous cleaning and deduplication through a rigorous pipeline of multiple stages to accomplish the best quality for model training, including language identification, URL-based filtering, metric-based cleaning, document refinement, and data deduplication. CulturaX is fully released to the public in HuggingFace to facilitate research and advancements in multilingual LLMs: \url{https://huggingface.co/datasets/uonlp/CulturaX}.

%We employ MinHash at the document level to achieve fuzzy deduplication for the datasets in different languages. Our data cleaning framework includes diverse criteria and threshold selections, guided by extensive data samples, ensuring comprehensive noise filtering in various aspects. 

\end{abstract}

\section{Introduction}

Large language models (LLMs) have fundamentally transformed research and applications of natural language processing (NLP), significantly advancing the state-of-the-art performance for numerous tasks and revealing new emergent abilities \cite{Brown2020LanguageMA,Wei2022EmergentAO}. Based on the transformer architecture \cite{Vaswani2017Attention}, three major variants of LLMs have been explored in the literature: the encoder-only models to encode input texts into representation vectors, e.g., BERT \cite{devlin-etal-2019-bert} and RoBERTa \cite{Liu2019RoBERTaAR}; the decoder-only models to generate texts, e.g., GPT \cite{Radford2019Language,Brown2020LanguageMA}; and the encoder-decoder models to perform sequence-to-sequence generation, e.g., BART \cite{lewis-etal-2020-bart} and T5 \cite{Raffel2020Xxploring}. The remarkable capabilities of LLMs have primarily been propelled by the ever-expanding scale of model sizes and training datasets, which have been deemed essential for achieving optimal performance by the scaling laws \cite{Hernandez2022ScalingLA}.  For instance, beginning with the BERT model, which had a mere few hundred million parameters \cite{devlin-etal-2019-bert}, recent GPT-based models have been expanded to encompass hundreds of billions of parameters \cite{Shoeybi2019MegatronLMTM,Scao2022BLOOMA1,Lieber2021Jurassic,Chowdhery2022PaLMSL}. Similarly, the training datasets for LLMs have grown exponentially, evolving from a modest 13GB of text data from Wikipedia and books used for BERT \cite{devlin-etal-2019-bert,Liu2019RoBERTaAR} to consume terabytes of data for the latest models, such as Falcon \cite{Penedo2023TheRD}, MPT \cite{MTP}, LLaMa \cite{Touvron2023LLaMAOA}, PolyLM \cite{Wei2023PolyLMAO} and ChatGPT\footnote{\url{https://openai.com/blog/chatgpt}}.

As the field keeps progressing rapidly, pre-trained LLMs have typically been released to the public to foster further research and advancements. These models are obtainable either through commercial APIs, as illustrated by ChatGPT and GPT-4, or via open-source initiatives, exemplified by Falcon and LLaMa. Nevertheless, in contrast to the public accessibility of LLMs, the training datasets that underpin the state-of-the-art models have mostly remained closely guarded secrets, even in the case of open-source LLMs such as BLOOM, LLaMa, MPT, and Falcon. For example, Falcon \cite{Penedo2023TheRD} and BLOOM \cite{Scao2022BLOOMA1} only provide a glimpse of their complete training data, whereas MPT's, LLaMa's and PolyLM's datasets \cite{Touvron2023LLaMAOA,Wei2023PolyLMAO} remain inaccessible to the public. On one hand, the lack of transparency has impeded in-depth analysis and comprehension of LLMs, hindering crucial research into attributing and addressing fundamental issues stemming from the training data, such as hallucinations, biases, and toxic content \cite{Tamkin2021UnderstandingTC,Weidinger2021EthicalAS,Kenton2021AlignmentOL,Bommasani2021OnTO}. On the other hand, concealing the training data restricts the development of LLMs to a select few stakeholders with ample resources, thereby constraining the democratization and benefits of the technology and exacerbating its biases within broader society.

To attain transparency and democratization for LLMs, it is thus crucial to create large-scale and high-quality datasets for training high-performing LLMs while ensuring their public accessibility to foster deeper research and advancements. In the realm of LLMs, high-quality training datasets are often crafted through the application of extensive data cleaning and deduplication processes, aimed at eliminating noisy and redundant content from vast text collections \cite{Allamanis2018TheAE,Penedo2023TheRD}. To this end, there have been recent efforts from the community to develop such open-source datasets for LLMs, such as RedPajama with 1.21T tokens \cite{together2023redpajama}, SlimPajama\footnote{\urlx{https://www.cerebras.net/blog/slimpajama-a-627b-token-cleaned-and-deduplicated-version-of-redpajama}} with 627B tokens, and AI2 Dolma\footnote{\urlx{https://blog.allenai.org/dolma-3-trillion-tokens-open-llm-corpus-9a0ff4b8da64}} with 3T tokens. However, most of the existing open-source datasets for LLMs are tailored for the English language, which hinders the utilization and performance of the resulting LLMs when applied to non-English languages, particularly those with limited linguistic resources \cite{Bang2023AMultitask,Lai2023ChatGPTBE}. This emphasis on English also restricts the capacity of open-source datasets to comprehensively tackle the research challenges and democratization concerns of LLMs across the diverse spectrum of over 7,000 languages spoken worldwide. 

Simultaneously, some multilingual datasets have been developed and made available, providing text data for multiple languages. Nevertheless, their quality and scale fall short of meeting the requirements for training high-performing LLMs. Specifically, the multilingual text dataset sourced from Wikipedia, while of high quality, is regarded as relatively small when it comes to training LLMs \cite{conneau-etal-2020-unsupervised}. The OSCAR datasets \cite{OrtizSuarezSagotRomary2019,ortiz-suarez-etal-2020-monolingual,AbadjiOrtizSuarezRomaryetal.2021,abadji-etal-2022-towards}\footnote{\url{https://oscar-project.org}} extract text data from CommonCrawl (CC) for more than 160 languages. However, these datasets lack document-level deduplication (i.e., removing similar documents in the dataset), leading to the inclusion of redundant information and impairing the performance of generative LLMs \cite{lee-etal-2022-deduplicating}. Similarly, the mC4 \cite{xue-etal-2021-mt5}, CCAligned \cite{conneau-etal-2020-unsupervised}, WikiMatrix \cite{schwenk-etal-2021-wikimatrix}, and ParaCrawl \cite{banon-etal-2020-paracrawl} datasets altogether support over 100 languages but suffers from less accurate language identification, introducing noise into the data \cite{kreutzer-etal-2022-quality}. These datasets are also not deduplicated at fuzzy and document levels, e.g., via MinHash \cite{Broder1997On}. Additionally, the CC100 dataset \cite{wenzek-etal-2020-ccnet,conneau-etal-2020-unsupervised}, employed in training the multilingual XLM-RoBERTa model across 100 languages, only considers the snapshots of CC in 2018, constraining its size and the availability of up-to-date information to train high-performing LLMs.

To address the aforementioned issues for open-source datasets, our work introduces a novel multilingual dataset, called CulturaX, for training LLMs in 167 languages. CulturaX merges the latest iteration of mC4 (version 3.1.0) with all available OSCAR corpora up to the current year, encompassing distributions 20.19, 21.09, 22.01, and 23.01. This amalgamation results in a large multilingual dataset, comprising 27 TB of text data with 6.3 trillion tokens and offering the most up-to-date data for LLM development. More than half of our dataset is dedicated to non-English languages to significantly boost the data size and enhance the feasibility of training models in multilingual scenarios. Importantly, CulturaX is extensively cleaned and deduplicated at the document level to produce the highest quality to train LLMs for multiple languages. In particular, our data cleaning process includes a comprehensive pipeline designed to eliminate low-quality data. This involves removing noisy text, non-linguistic content, toxic data, incorrect language identification, and more. Our data cleaning pipeline employs a variant of the Interquartile Range (IQR) method \cite{Dekking2007AMI} to select appropriate thresholds for various dataset metrics (e.g., stopword ratios, data perplexity, and language identification scores), which can be used to filter noisy outliers for the dataset. As such, we leverage the percentiles of the distributions computed over large samples of data to effectively guide the threshold selection process for each filtering metric and language. Finally, we perform extensive deduplication for the data of the languages within our datasets based on the near deduplication method MinHashLSH \cite{Broder1997On,Leskovec2020} and URLs, leading to high-quality data to train multilingual LLMs. Our dataset will be fully available to the public to promote further research and development for multilingual learning. To our knowledge, CulturaX is the largest open-source multilingual dataset to date that is deeply cleaned and deduplicated for LLM and NLP applications.

\section{Multilingual Dataset Creation}

To develop a multilingual public dataset for LLMs, our strategy is to combine mC4 \cite{xue-etal-2021-mt5} and OSCAR \cite{OrtizSuarezSagotRomary2019,AbadjiOrtizSuarezRomaryetal.2021,abadji-etal-2022-towards}, two largest multilingual datasets at our disposal. We then process the data with an extensive pipeline, involving two major steps of cleaning and deduplication, to produce an enormous and high-quality dataset for multilingual LLMs.

{\bf mC4} is a multilingual document-level dataset, originally created to train the multilingual encoder-decoder model mT5 \cite{xue-etal-2021-mt5} for 101 languages. This dataset is extracted from 71 monthly snapshots from CC by removing pages with less than three long lines (line length filter), pages with bad words, and duplicated lines across documents. Language identification for the pages in mC4 is done by the {\tt cld3} tool \cite{botha-etal-2017-natural}\footnote{\url{https://github.com/google/cld3}}, which is a small feed-forward network \cite{xue-etal-2021-mt5}. Any pages with a language confidence below 0.95\% are excluded. mC4 is deduplicated with exact match at the document level; however, fuzzy document-level deduplication is not performed. We utilize the latest version of mC4 (version 3.1.0)\footnote{\url{https://huggingface.co/datasets/mc4}} prepared by AllenAI in this work.

\begin{figure}
    \centering
    \includegraphics[width=\linewidth]{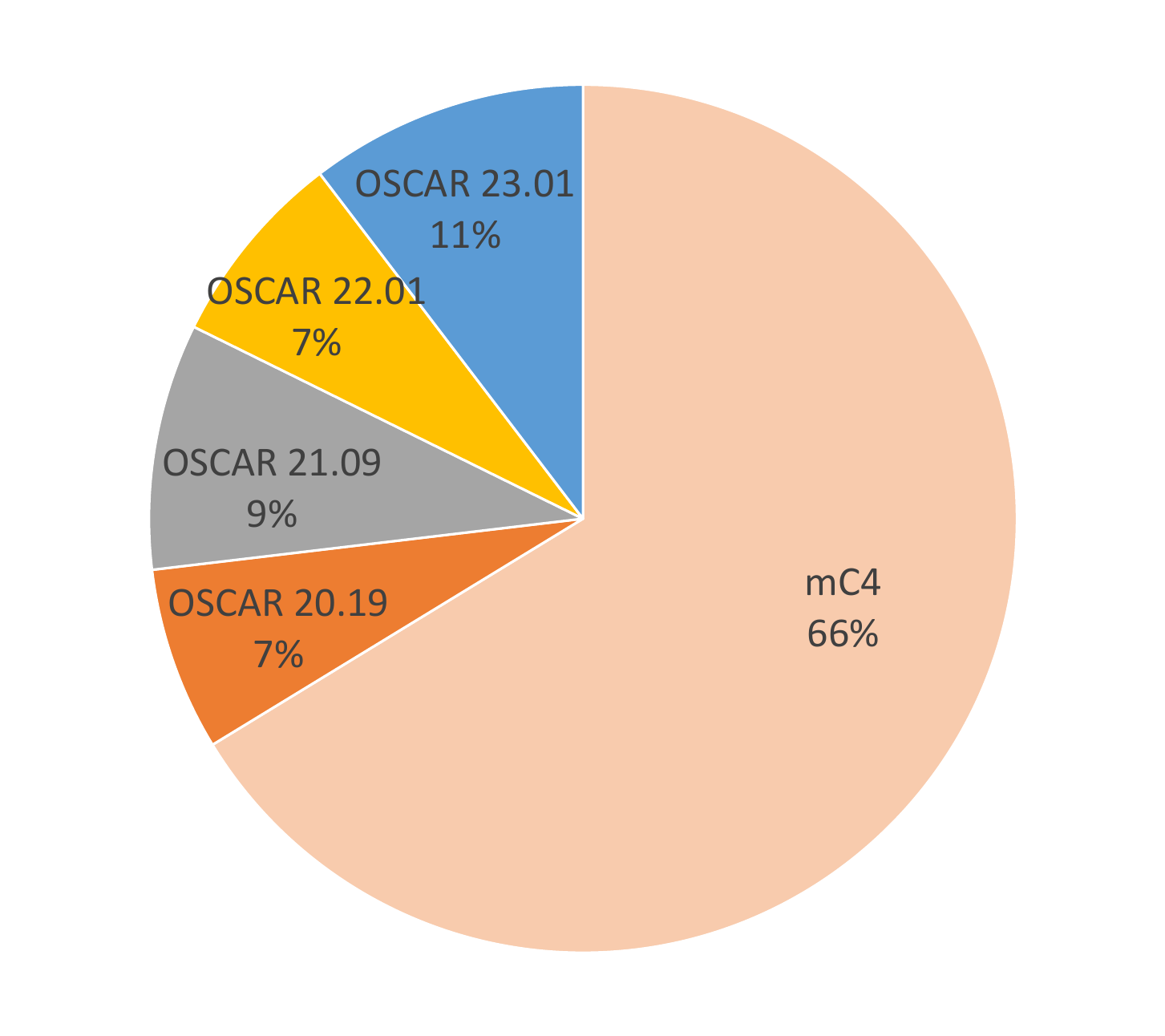}
    \caption{Distribution of document counts from mC4 and OSCAR in our initial dataset.}
    \label{fig:portion}
\end{figure}

A notable aspect of our dataset pertains to the web-based origin of our selected datasets, mC4 and OSCAR, extracted from CC. This differs from certain previous work \cite{Radford2019Language,MTP,Touvron2023LLaMAOA} that has also relied on curated datasets like The Pile \cite{Gao2020ThePA} and BookCorpus \cite{Zhu2015AligningBA} to train LLMs, presuming their higher overall quality. However, in the context of multilingual settings, we argue that web-scraped datasets can be a more suitable approach, as curated datasets of superior quality might not be available for various languages. Our strategy of using web-scraped data facilitates efficient data collection across multiple languages, contributing to enhanced training data scales. Furthermore, recent studies have demonstrated the effectiveness of cleaning web-scraped data to yield state-of-the-art LLMs \cite{Raffel2020Xxploring,falcon40b}. In total, the combination of mC4 and OSCAR provides us 13.5B documents for further processing. Figure \ref{fig:portion} illustrates the distribution of the document counts for mC4 and the four available versions of OSCAR in our initial dataset.

%One notable detail in our dataset concerns the web-based nature of our selected datasets mC4 and OSCAR from CC. This is in contrast to some previous work \cite{Radford2019Language,MTP,Touvron2023LLaMAOA} that also exploits curated datasets to train LLMs, such as The Pile \cite{Gao2020ThePA} and BookCorpus \cite{Zhu2015AligningBA}, assuming their overall higher quality. However, in the multilingual settings, we argue web-scrape datasets can be a more suitable approach as curated datasets of high quality might not exist for different languages. Our strategy with web-scrape data can allow convenient data collection for multiple languages to efficiently increase the data scales. It has also been demonstrated in some recent work that web-scrape data can be cleaned well to produce state-of-the-art LLMs \cite{Raffel2020Xxploring,falcon40b}.

%\subsection{Data Collection}

\subsection{Data Cleaning}

Given the combination of the mC4 and OSCAR datasets, we first perform a comprehensive data cleaning procedure to remove noisy and bad content from the data, including language identification, ULR-based filtering, metric-based cleaning, and document refinement.

%Table \ref{tab:stats} provides statistics for our dataset across different stages of the cleaning process.

{\bf Language Identification}: A particular issue concerns the use of two different language identification tools, i.e., {\tt cld3} and FastText, for mC4 and OSCAR (respectively). It has been shown in previous studies that {\tt cld3} is significantly worse than FastText, causing substantially more language detection errors for mC4 \cite{kreutzer-etal-2022-quality}. In fact, compared to several other language detectors, FastText has demonstrated state-of-the-art performance over benchmark datasets\footnote{\url{https://modelpredict.com/language-identification-survey}}. To this end, our first data cleaning step involves applying FastText to re-predict the languages for the documents in mC4. Documents whose predicted languages are different from the provided ones in mC4 will be removed from the dataset. The rationale is to avoid documents that are confusing for the language detectors {\tt cld3} and FastText, thus potentially introducing noise for the data. Finally, to ensure the highest quality, we remove data for any language found in mC4 but not supported by FastText.

%Finally, we also discard any language that is present in mC4, but not supported by FastText for consistency.

{\bf URL-based Filtering}: In the next step, we aim to eliminate pages from the known toxic and harmful sources to reduce relevant risks from our data. In particular, we leverage the latest UT1 blacklist of URLs and domains provided by the University of Toulouse to support Internet use regulation for administrators at schools. This list involves sites from different topics, including pornography, grumbling, and hacking, that should be discarded for LLM training. Updated twice to thrice per week, the blacklist involves more than 3.7M records that are contributed by both human and robots (e.g., search engines, known addresses and indexes) \cite{abadji-etal-2022-towards}. As such, we remove any page from our dataset whose associated URL matches a site in the blacklist. This step is helpful for our dataset as the blacklist is not employed before for the mC4 dataset. In addition, although OSCAR has already used this blacklist for data cleaning, our approach incorporates the most up-to-date information from the list, which might not be available for the current distributions of OSCAR.

{\bf Metric-based Cleaning}: To enhance the dataset's quality, motivated by the data processing pipeline from the BigScience's ROOTS corpus for BLOOM \cite{laurenon2022the,Scao2022BLOOMA1}, we further utilize the distributions for various dataset metrics to identify and filter outlying documents. Each metric provides a singular value for every document within the dataset, quantifying specific attributes such as {\it number\_words}, {\it stopword\_ratios}, and {\it perplexity\_score} for each document. For each metric and its range of possible values within the dataset, a threshold will be determined to partition the range into two zones: a normal range and an abnormal range. The abnormal range is designated for documents exhibiting metric values significantly deviating from the norm, classifying them as outliers/noises, and consequently, these outliers are removed from our dataset. As such, we employ a comprehensive array of dataset metrics, which will be collectively employed to refine our dataset, as outlined below:
\begin{itemize}[noitemsep]
    \item Number of words
    \item Character repetition ratio
    \item Word repetition ratio
    \item Special character ratio
    \item Stop word ratio
    \item Flagged word ratio
    \item Language identification confidence
    \item Perplexity score
    \item Document length (number of characters)
    \item Number of lines
    \item Short line length ratio
    \item Short line ratio
\end{itemize}

The last four metrics are suggested by the OSCAR dataset while the others are inherited from the BigScience ROOTS corpus's pipeline to process OSCAR data. For the perplexity score, following the BigScience ROOTS corpus, we train a SentencePiece tokenizer \cite{kudo-2018-subword} and 5-gram Kneser-Ney language models as provided in the KenLM library \cite{heafield-2011-kenlm} using the 20230501 dumps of Wikipedia. Documents displaying high perplexity scores based on these KenLM models are considered notably different from Wikipedia articles. This indicates a level of noise that will be excluded from our dataset \cite{wenzek-etal-2020-ccnet}. The tokenizer will also be used to obtain the number of words/tokens in the documents for our metrics. We publicly release our KenLM models in HuggingFace\footnote{\url{https://huggingface.co/uonlp/kenlm}} to faciliate future exploration.

Repeated information (e.g., words, paragraphs) can appear in the web-curated data due to crawling errors and low-quality sources, causing detrimental consequences for training LLMs \cite{Holtzman2019TheCC}. The character and word repetition ratios are thus designed to avoid documents with excessively repeated information. High frequencies of special characters, stop words, or flagged words can indicate noisy and low-quality documents. We thus utilize the stop word and flagged word lists for different languages to compute their ratios for document removal. In addition to the stop word and flagged word lists provided by BigScience ROOTS for their 13 languages, we further collect dictionaries for these types of words for other languages. We prioritize the lists that have been shared on personal GitHub accounts for various languages, as these are often crafted by native speakers and exhibit higher quality. Moreover, lower language identification confidence might also suggest noisy language structures for the data. For each document in the dataset, we thus obtain a language identification confidence via the probability that FastText assigns to its corresponding language to aid data filtering. Finally, for the short line-based criteria, we implement a threshold of 100 characters to classify lines as short, as used by OSCAR. Documents with excessive occurrence of short lines will not be retained in our dataset.

{\bf Threshold Selection}: Given the set of dataset metrics, an important question concerns the selection of appropriate thresholds for each metric and language to generate high-quality multilingual data. In the BigScience ROOTS project \cite{laurenon2022the}, this selection process is carried out by native speakers of 13 languages. The resulting thresholds are employed for the rest of their 46 languages. The project offers a visualization interface that indexes a sample of a few thousand documents per language, enabling users to monitor data statistics as they adjust thresholds for the metrics. However, this process cannot be easily extended to different languages due to the requirement of experienced native speakers, which incurs significant costs. Furthermore, the limited sample sizes hinder the representativeness of the chosen thresholds for the full datasets. In our analysis, we observe that some selected thresholds for certain languages within BigScience ROOTS almost fall outside the value ranges for the entire dataset, leading to the deactivation of the corresponding metrics. 

To address these issues, we leverage a variant of the Interquartile Range (IQR) method \cite{Dekking2007AMI} to select appropriate thresholds for the filtering metrics for our dataset. For each metric and language, we generate a distribution of its possible values across the entire dataset for the language. There is an exception for languages with substantial amounts of data, such as Spanish and Russian, where only 25\% of the data is used to calculate these distributions. Afterward, we compute the $Q_1$-th and $Q_3$-th percentiles of the distribution ($Q_1 < Q3$) and use them for the thresholds for our filtering metrics. In particular, the lower $Q_1$-th percentile will be chosen for the metrics that favor high values (e.g., language identification confidence), while metrics favoring low values (e.g., perplexity scores and document length) will utilize the upper $Q_3$-th percentile. We investigate different values for $(Q_1,Q_3)$, considering $(25, 75)$, $(20, 80)$, $(15, 85)$, $(10, 90)$, and $(5, 95)$. The selection of $Q_1 = 10$ and $Q_2 = 90$ has achieved the best data quality for a sample of languages in our examination.

It is worth emphasizing that the utilization of percentiles for threshold selection enables our approach to efficiently draw upon more extensive data samples for each language compared to those employed in the BigScience ROOTS project. This results in more reliable thresholds for the full datasets over different languages. Specifically, concerning the large languages where only a 25\% data sample is employed to compute the value distribution for a metric, we observe that the proportion of discarded data to the entire dataset closely aligns with that of the data sample when applying the same selected filtering threshold. This underscores the representativeness of the thresholds selected through our methodology. Finally, once the thresholds for the metrics in a given language have been determined, we will eliminate any document that surpasses a metric's threshold and enters the unfavorable range of the data.

{\bf Document Refinement}: The previous cleaning steps are done at the dataset level, aiming to remove low-quality documents from the dataset. In this step, we further clean the retained documents to improve the quality. It is important to note that our prior metric-based filtering step plays a vital role in eliminating highly noisy documents, which, in turn, streamlines the process of developing effective document cleaning rules during this step. Notably, since the documents from mC4 and OSCAR are extracted from HTML pages crawled from the Internet, a significant portion of them may carry crawling and extraction errors, including long JavaScript lines and extraneous content. Consequently, filtering out these documents greatly simplifies our task of designing rules to clean the documents within our dataset.

As such, for each document, we eliminate its noisy or irrelevant portions via a series of operations. First, we remove any short lines located at the end of each document, as these lines typically contain footer details or unhelpful information from the websites. Second, we eliminate the lines that contain words from our list of JavaScript (JS) keywords (e.g., ``{\tt <script}'') to avoid irrelevant and non-linguistic information. Here, we exclusively remove JS lines if the document contains just one line with JS keywords, and this particular line must also feature at least two different types of JS keywords. We adopt this approach as documents with more than two JS lines are likely coding tutorials in our data, which should be preserved to improve diversity. In addition, certain JS keywords are used in natural language, e.g., ``{\tt var}''. By requiring at least two different types of JS keywords, we reduce the risk of inadvertently omitting helpful content and disrupting the document's structure.

\subsection{Data Deduplication}

Despite thorough data cleaning, the remaining dataset might still contain a substantial amount of repeated data due to various reasons, including information being reposted on the web, multiple references to the same articles, boilerplate content, and plagiarism. The duplicated data can thus cause memorization and significantly hinder generalization for LLMs \cite{lee-etal-2022-deduplicating,Hernandez2022ScalingLA}. Although expensive, data deduplication is thus considered as a crucial step to guarantee the highest quality of data for training LLMs. To this end, we undertake a comprehensive deduplication procedure for our dataset, utilizing MinHash \cite{Broder1997On} and URLs. This deduplication process is carried out independently for each language. Furthermore, we restrict deduplication to languages that retain over 100K documents following our data cleaning procedures (i.e., $51.5$\% of our languages), aiming to promote smaller languages within our dataset.

%Exact Match \cite{lee-etal-2022-deduplicating},

%we perform an extensive deduplication process for our dataset, utilizing MinHash \cite{Broder1997On}, Exact Match \cite{lee-etal-2022-deduplicating}, and URLs. The deduplication is done separately for each language. In addition, we only deduplicate the data for languages that have more than 100K documents after our cleaning process to promote small languages in our dataset.

\newcommand{\mytt}[2]{\multicolumn{#1}{c}{\textbf{#2}}}
\newcommand{\mymtr}[2]{\multirow{#1}{*}{\multicolumn{1}{c}{\textbf{#2}}}}

\begin{table*}[htbp]
  \centering
\resizebox{\textwidth}{!}{
    \begin{tabular}{llrrrrrrrr}
    %{clrrrrr@{\extracolsep{6pt}}rr@{}r}
    \hline
      %\mymtr{3}{Code}    & \mymtr{3}{Language}     & \mytt{6}{\#Documents (M)} & \mytt{2}{\#Tokens}   \\ 
      \multirow{3}{*}{\bf Code} & \multirow{3}{*}{\bf Language} & \multicolumn{5}{c}{\bf \#Documents (M)} & & \multicolumn{2}{c}{\bf \#Tokens} \\
      \cline{3-7}
      \cline{9-10}
      & & \multirow{2}{*}{\centering \bf Initial} & \multicolumn{1}{c}{\bf URL} &  \multicolumn{1}{c}{\bf Metric} &  \multicolumn{1}{c}{\bf MinHash} &  \multicolumn{1}{c}{\bf URL} & \multicolumn{1}{c}{\bf Filtering} & \multicolumn{1}{c}{\bf (B)} & \multicolumn{1}{c}{\bf (\%)} \\
      & & & \multicolumn{1}{c}{\bf Filtering} &  \multicolumn{1}{c}{\bf Filtering} &  \multicolumn{1}{c}{\bf Dedup} &  \multicolumn{1}{c}{\bf Dedup} & \multicolumn{1}{c}{\bf Rate (\%)} & & \\ \hline
%      &  & \centering {\bf Initial} & \centering {\bf URL} & \centering {\bf Metric} & \centering {\bf MinHash} & \centering {\bf URL} & \centering {\bf Filtering} & \centering {\bf (B)} & \centering {\bf (\%)}  \\
%      &  &  & & \centering {\bf Filtering} & \centering {\bf Filtering} & \centering {\bf Dedup} & \centering {\bf Dedup} & \centering {\bf Rate (\%)} & & \\ \hline
      %& & & &\centering {\bf Filtering}&&&&& \\
    en    & English & 5783.24 & 5766.08 & 3586.85 & 3308.30 & 3241.07 & 43.96 & 2846.97 & 45.13 \\
    ru    & Russian & 1431.35 & 1429.05 & 922.34 & 845.64 & 799.31 & 44.16 & 737.20 & 11.69 \\
    es    & Spanish & 844.48 & 842.75 & 530.01 & 479.65 & 450.94 & 46.60  & 373.85 & 5.93 \\
    de    & German & 863.18 & 861.46 & 515.83 & 447.06 & 420.02 & 51.34 & 357.03 & 5.66 \\
    fr    & French & 711.64 & 709.48 & 439.69 & 387.37 & 363.75 & 48.89 & 319.33 & 5.06 \\
    zh    & Chinese & 444.37 & 444.03 & 258.35 & 222.37 & 218.62 & 50.80  & 227.06 & 3.60 \\
    it    & Italian & 406.87 & 406.04 & 254.72 & 226.42 & 211.31 & 48.06 & 165.45 & 2.62 \\
    pt    & Portuguese & 347.47 & 346.76 & 217.21 & 200.11 & 190.29 & 45.24 & 136.94 & 2.17 \\
    pl    & Polish & 270.12 & 269.73 & 170.86 & 151.71 & 142.17 & 47.37 & 117.27 & 1.86 \\
    ja    & Japanese & 247.67 & 247.19 & 137.88 & 114.64 & 111.19 & 55.11 & 107.87 & 1.71 \\
    vi    & Vietnamese & 182.88 & 182.72 & 118.67 & 108.77 & 102.41 & 44.00    & 98.45 & 1.56 \\
    nl    & Dutch & 238.92 & 238.56 & 148.19 & 125.51 & 117.39 & 50.87 & 80.03 & 1.27 \\
    ar    & Arabic & 132.88 & 132.65 & 84.84 & 77.65 & 74.03 & 44.29 & 69.35 & 1.10 \\
    tr    & Turkish & 183.65 & 183.47 & 109.94 & 99.18 & 94.21 & 48.70  & 64.29 & 1.02 \\
    cs    & Czech & 136.91 & 136.44 & 80.38 & 69.01 & 65.35 & 52.27 & 56.91 & 0.90 \\
    fa    & Persian & 118.55 & 118.50 & 70.26 & 62.42 & 59.53 & 49.78 & 45.95 & 0.73 \\
    hu    & Hungarian & 88.59 & 88.21 & 53.29 & 46.89 & 44.13 & 50.19 & 43.42 & 0.69 \\
    el    & Greek & 100.77 & 100.68 & 61.43 & 54.33 & 51.43 & 48.96 & 43.15 & 0.68 \\
    ro    & Romanian & 89.37 & 89.25 & 45.99 & 42.8  & 40.33 & 54.87 & 39.65 & 0.63 \\
    sv    & Swedish & 103.04 & 102.76 & 58.67 & 52.09 & 49.71 & 51.76 & 38.49 & 0.61 \\
    uk    & Ukrainian & 81.50  & 81.44 & 50.95 & 47.12 & 44.74 & 45.10  & 38.23 & 0.61 \\
    fi    & Finnish & 59.85 & 59.80  & 36.69 & 32.15 & 30.47 & 49.09 & 28.93 & 0.46 \\
    ko    & Korean & 46.09 & 45.85 & 25.19 & 21.17 & 20.56 & 55.39 & 24.77 & 0.39 \\
    da    & Danish & 53.16 & 52.99 & 28.67 & 26.48 & 25.43 & 52.16 & 22.92 & 0.36 \\
    bg    & Bulgarian & 47.01 & 46.90  & 28.09 & 25.45 & 24.13 & 48.67 & 22.92 & 0.36 \\
    no    & Norwegian & 40.07 & 40.01 & 20.69 & 19.49 & 18.91 & 52.81 & 18.43 & 0.29 \\
    hi    & Hindi & 35.59 & 35.50  & 22.01 & 20.77 & 19.67 & 44.73 & 16.79 & 0.27 \\
    sk    & Slovak & 40.13 & 39.95 & 22.20  & 19.56 & 18.58 & 53.70  & 16.44 & 0.26 \\
    th    & Thai  & 49.04 & 48.96 & 26.20  & 21.93 & 20.96 & 57.26 & 15.72 & 0.25 \\
    lt    & Lithuanian & 27.08 & 27.01 & 15.87 & 14.25 & 13.34 & 50.74 & 14.25 & 0.23 \\
    ca    & Catalan & 31.13 & 31.12 & 18.99 & 16.46 & 15.53 & 50.11 & 12.53 & 0.20 \\
    id    & Indonesian & 48.08 & 48.05 & 25.79 & 23.74 & 23.25 & 51.64 & 12.06 & 0.19 \\
    bn    & Bangla & 20.90  & 20.85 & 13.82 & 13.22 & 12.44 & 40.48 & 9.57  & 0.15 \\
    et    & Estonian & 16.20  & 16.15 & 9.69  & 8.45  & 8.00     & 50.62 & 8.81  & 0.14 \\
    sl    & Slovenian & 15.46 & 15.39 & 8.00     & 7.60   & 7.34  & 52.52 & 8.01  & 0.13 \\
    lv    & Latvian & 14.14 & 14.09 & 8.37  & 7.48  & 7.14  & 49.50  & 7.85  & 0.12 \\
    he    & Hebrew & 10.78 & 10.77 & 5.90   & 4.77  & 4.65  & 56.86 & 4.94  & 0.08 \\
    sr    & Serbian & 7.80   & 7.75  & 4.80   & 4.25  & 4.05  & 48.08 & 4.62  & 0.07 \\
    ta    & Tamil & 8.77  & 8.75  & 5.27  & 4.94  & 4.73  & 46.07 & 4.38  & 0.07 \\
    sq    & Albanian & 9.40   & 9.38  & 5.96  & 5.04   & 5.21  & 44.57 & 3.65  & 0.06 \\
    az    & Azerbaijani & 9.66  & 9.65  & 5.73  & 5.24  & 5.08  & 47.41 & 3.51  & 0.06 \\ \hline
    \multicolumn{2}{l}{\bf Total (42 languages)} & \bf 13397.79 & \bf 13366.17	& \bf 8254.28 & \bf 7471.48 & \bf 7181.40 & \bf 46.40 & \bf 6267.99 & \bf 99.37 \\
    \hline
    \multicolumn{2}{l}{\bf Total (167 languages)} & \bf 13506.76 & \bf13474.94 & \bf 8308.74 & \bf 7521.23 & \bf 7228.91 & \bf 46.48 & \bf 6308.42 & \bf 100.00 \\
    \hline
    \end{tabular}%
}
\caption{Data statistics for 42 languages with the percentages of tokens greater than 0.05\% in our dataset. Columns grouped with the ``\#Documents (M)'' label indicate the number of documents for each language after the corresponding cleaning and reduplication steps. The token counts are based on our final dataset (i.e., after all the cleaning and deduplication steps).}
  \label{tab:stats}%
\end{table*}%

{\bf MinHash Deduplication}: For each language's dataset, we first apply the MinHashLSH method \cite{Leskovec2020} to filter similar documents in the dataset. MinHashLSH is a near deduplication technique based on MinHash \cite{Broder1997On} with multiple hash functions for $n$-grams and the Jaccard similarity. Locality-Sensitive Hashing (LSH) is incorporated to improve efficiency by focusing on document pairs that are most likely similar. We leverage a variant of the Spark implementation of MinHashLSH in the {\tt text-dedup} repo\footnote{\url{https://github.com/ChenghaoMou/text-dedup/tree/main}}, employing $5$-grams and a threshold of $0.8$ to determine similar documents for the Jaccard similarity. Running MinHashLSH for each language's dataset, especially for languages with the largest data volumes like English, Russian, Spanish, and Chinese, represents the most computationally expensive operation in our dataset creation effort.

%9,000 hashes for each document to deduplicate our multilingual datasets

{\bf URL-based Deduplication}: Finally, we eliminate all documents that share identical URLs with other documents in the dataset. This step is necessary to address situations where various versions of the same articles are linked to identical URLs but have been updated or modified during the publication process, effectively bypassing the near deduplication step. Some URLs for the articles in CC might only display their general domains due to crawling errors. To enhance accuracy, we refrain from removing URLs that only include their general domains.

%only have their general domains due to the crawling errors. To improve accuracy, we do not remove URLs that only have the general domains.

%This step is to account for different versions of the same articles that are assigned to the same URLs, but updated/modified during the publication process, thus bypassing the near deduplication step.

%We perform URL-based deduplication first as it can be done efficiently to quickly reduce the data size. A smaller dataset can be more manageable for the subsequent deduplication steps.

%{\bf Exact Match Deduplication}: URL- and MinHash-based deduplications compare and discard data at dataset level. In the last stage, we perform deduplication at the document level, eliminating identical substrings found in a document when compared to others. In particular, we match substrings with 50 consecutive tokens for removal from our dataset. We also use the implementation of Extract Match Deduplication in the {\tt text-dedup} repo for this step. Although exact match deduplication can modify the input document, it has been shown to produce cleaner data to improve LLM performance \cite{lee-etal-2022-deduplicating,Penedo2023TheRD}.

We utilize 600 AWS c5.24xlarge EC2 instances to preprocess and deduplicate our multilingual dataset. Each instance is equipped with 96 CPU cores, 192GB of memory, and 1TB of disk space. The disk space can be used to replace memory when necessary (e.g., for data deduplication).

%\subsection{Data Delivery}

\section{Data Analysis and Experiments}

After completing all the cleaning and deduplication steps, our ultimate dataset comprises 6.3 trillion tokens spanning 167 languages. Table \ref{tab:stats} provides an overview of the number of documents and tokens for the top 42 languages in CulturaX following each processing stage. As can be seen, our data-cleaning pipeline can substantially reduce the number of documents in the original mC4 and OSCAR datasets for each language. The total number of removed documents accounts for 46.48\% of our initial documents, suggesting the the effectiveness of our approaches to filter noisy information for multilingual datasets.

\section{Related Work}

%Even though research and development efforts for LLMs have seen significant growth, the community has paid relatively less attention to the creation of datasets for training these models. 

Compared to other NLP tasks, language models can be trained with unlabeled data, enabling efficient data collection to produce gigantic scales for the training data. There are two primary types of data commonly used for training LLMs: curated data and web crawl data. Curated data typically consists of well-written and well-formatted text from targeted sources and domains, e.g., Wikipedia articles, books, newswire articles, and scientific papers, as used for the ``The Pile'' \cite{Gao2020ThePA} and ``BookCorpus'' \cite{Zhu2015AligningBA} datasets. In contrast, web crawl data encompasses text gathered from a wide array of sources across the internet, varying significantly in terms of format and writing styles, e.g., blogs, social media posts, news articles, and advertisements. CommonCrawl (CC) is a widely-used web crawl repository that has collected petabytes of data over the Internet for 12 years. To this end, curated data is frequently considered to possess higher quality, which has resulted in its preference for training early LLMs, e.g., BERT \cite{devlin-etal-2019-bert} and GPT-2 \cite{Radford2019Language}. However, as the demand for larger models has grown, web crawl data has gained more attention as it contributes a substantial portion to the training data of recent LLMs, e.g., RoBERTa \cite{Liu2019RoBERTaAR}, BART \cite{lewis-etal-2020-bart}, T5 \cite{Raffel2020Xxploring}, GPT-3 \cite{Rae2021ScalingLM}, LLaMa \cite{Touvron2023LLaMAOA}, MPT \cite{MTP}, and Falcon \cite{falcon40b}. As such, different extractions of CC has been produced to train such LLMs, including C4 \cite{Raffel2020Xxploring}, CC-News \cite{ccnews}, and STORIES \cite{Trinh2018ASM}.

Regarding the accessibility of training data, datasets used to train early LLMs are often made available to the public \cite{devlin-etal-2019-bert,Raffel2020Xxploring}. However, in the case of the most recent state-of-the-art (SOTA) generative LLMs, their training datasets are not released fully, potentially due to commercial interests. This applies not only to proprietary models like ChatGPT and GPT-4 but also to models that claim to be open-source models such as LLaMa, MPT, Falcon, and BLOOM \cite{Scao2022BLOOMA1}. To address the transparency issue with existing LLMs, recent efforts have been made to replicate and release the training datasets for the state-of-the-art LLMs, i.e., RedPajama \cite{together2023redpajama}, SlimPajama, and AI2 Dolma. The key distinctions for these datasets concern their large-scale text data that has been meticulously cleaned and document-level deduplicated to ensure high quality for training LLMs. Nonetheless, a common drawback of these open-source datasets is that they remain predominantly focused on English data, offering limited data for other languages. 

To obtain a multilingual large-scale dataset for training LLMs, it is more convenient to exploit web-scrape datasets such as CC to enable efficient data collection with up-to-date information in multiple languages. In addition, to ensure high quality for high-performing LLMs, it is necessary to extensively clean and deduplicate the multilingual data to avoid noisy and irrelevant content, e.g., low-quality machine-generated text and adult content \cite{Trinh2018ASM,kreutzer-etal-2022-quality,Raffel2020Xxploring}. As such, a typical data processing pipeline to generate high-quality datasets can involve multiple steps, as demonstrated by FastText \cite{Joulin2016FastTextzipCT}, CC-Net \cite{wenzek-etal-2020-ccnet}, the BigScience ROOTS corpus for the BLOOM models \cite{laurenon2022the,Scao2022BLOOMA1}, the RefinedWeb dataset for the Falcon model \cite{Penedo2023TheRD,falcon40b}, and the dataset to train the LLaMa models \cite{Touvron2023LLaMAOA}. The first step necessitates in such pipelines language identification to appropriately assign data to their corresponding languages \cite{Joulin2016FastTextzipCT}. The next steps features various dataset-specific rules and heuristics to filter undesirable content according to the ratios of special characters, short lines, bad words, among others \cite{grave-etal-2018-learning,laurenon2022the}. The data can also be filtered via lightweight models, e.g., via the KenLM language models \cite{heafield-2011-kenlm}, to avoid noisy documents \cite{wenzek-etal-2020-ccnet}. Finally, data deduplication should be performed to remove similar or repeated information \cite{laurenon2022the,Penedo2023TheRD}. An important step in this regard involves fuzzy deduplication at document level, e.g., via MinHash \cite{Broder1997On}, to eliminate similar documents, thus mitigating memorization and improving the generalization for resulting LLMs  \cite{lee-etal-2022-deduplicating}.

%The first step necessitates in such pipelines language identification to appropriately assign data to their corresponding languages that can be done by language detectors such as FastText \cite{Joulin2016FastTextzipCT} and {\tt cld3} \cite{botha-etal-2017-natural}. The next steps features various dataset-specific rules and heuristics to filter undesirable content according to the ratios of special characters, short lines, bad words, among others \cite{grave-etal-2018-learning,laurenon2022the}. The data can also be filtered via lightweight models, e.g., via the KenLM language models \cite{heafield-2011-kenlm}, to avoid noisy documents \cite{wenzek-etal-2020-ccnet}. Finally, data deduplication should be performed to remove similar or repeated information \cite{laurenon2022the,Penedo2023TheRD}. An important step in this regard involves fuzzy dedupication at document level, e.g., via MinHash \cite{Broder1997On}, to eliminate similar documents, thus mitigating memorization and improving the generalization for resulting LLMs  \cite{lee-etal-2022-deduplicating}.

To this end, while there are multilingual open-source datasets with text data in multiple languages, such as mC4 \cite{xue-etal-2021-mt5}, OSCAR \cite{OrtizSuarezSagotRomary2019}, CC100 \cite{wenzek-etal-2020-ccnet,conneau-etal-2020-unsupervised}, and the BigScience ROOT corpus \cite{laurenon2022the}, their quality and scale do not meet the requirements for effectively training LLMs, particularly generative models such as GPT. For example, as highlighted in the introduction, both mC4 and OSCAR lack fuzzy deduplication for the data at the document level. mC4 also suffers from its poorer language identification due to the use of {\tt cld3}. BigScience ROOTS only provides a small sample data for 46 languages while CC100 does not have information beyond 2018. Our dataset CulturaX thus comprehensively addresses the issues for the existing datasets, offering a multilingual, open-source, and large-scale dataset with readily usable and high-quality data to train LLMs.

\section{Conclusion}

We present CulturaX, a novel multilingual dataset with text data for 167 languages. Our dataset is cleaned and deduplicated via a comprehensive pipeline, producing 6.3 trillion tokens. CulturaX is thus a large-scale and high-quality dataset, which can be readily used to train high-performing LLMs for multiple languages. Our data is openly accessible to the public to promote further research and applications of multilingual learning.

\bibliography{anthology,custom}

\begin{thebibliography}{52}
\expandafter\ifx\csname natexlab\endcsname\relax\def\natexlab#1{#1}\fi

\bibitem[{Abadji et~al.(2022)Abadji, Ortiz~Suarez, Romary, and
  Sagot}]{abadji-etal-2022-towards}
Julien Abadji, Pedro Ortiz~Suarez, Laurent Romary, and Beno{\^\i}t Sagot. 2022.
\newblock \href {https://aclanthology.org/2022.lrec-1.463} {Towards a cleaner
  document-oriented multilingual crawled corpus}.
\newblock In \emph{Proceedings of the Thirteenth Language Resources and
  Evaluation Conference}, pages 4344--4355, Marseille, France. European
  Language Resources Association.

\bibitem[{Abadji et~al.(2021)Abadji, Su{\'a}rez, Romary, and
  Sagot}]{AbadjiOrtizSuarezRomaryetal.2021}
Julien Abadji, Pedro Javier~Ortiz Su{\'a}rez, Laurent Romary, and Beno{\^i}t
  Sagot. 2021.
\newblock Ungoliant: An optimized pipeline for the generation of a very
  large-scale multilingual web corpus.
\newblock In \emph{Proceedings of the Workshop on Challenges in the Management
  of Large Corpora (CMLC-9) 2021. Limerick, 12 July 2021 (Online-Event)}.

\bibitem[{Allamanis(2018)}]{Allamanis2018TheAE}
Miltiadis Allamanis. 2018.
\newblock The adverse effects of code duplication in machine learning models of
  code.
\newblock \emph{Proceedings of the 2019 ACM SIGPLAN International Symposium on
  New Ideas, New Paradigms, and Reflections on Programming and Software}.

\bibitem[{Almazrouei et~al.(2023)Almazrouei, Alobeidli, and Alshamsi~et
  al.}]{falcon40b}
Ebtesam Almazrouei, Hamza Alobeidli, and Abdulaziz Alshamsi~et al. 2023.
\newblock {Falcon-40B}: an open large language model with state-of-the-art
  performance.

\bibitem[{Bang et~al.(2023)Bang, Cahyawijaya, Lee, Dai, Su, Wilie, Lovenia, Ji,
  Yu, Chung, Do, Xu, and Fung}]{Bang2023AMultitask}
Yejin Bang, Samuel Cahyawijaya, Nayeon Lee, Wenliang Dai, Dan Su, Bryan Wilie,
  Holy Lovenia, Ziwei Ji, Tiezheng Yu, Willy Chung, Quyet~V. Do, Yan Xu, and
  Pascale Fung. 2023.
\newblock A multitask, multilingual, multimodal evaluation of chatgpt on
  reasoning, hallucination, and interactivity.
\newblock \emph{ArXiv}, abs/2302.04023.

\bibitem[{Ba{\~n}{\'o}n et~al.(2020)Ba{\~n}{\'o}n, Chen, Haddow, Heafield,
  Hoang, Espl{\`a}-Gomis, Forcada, Kamran, Kirefu, Koehn, Ortiz~Rojas,
  Pla~Sempere, Ram{\'\i}rez-S{\'a}nchez, Sarr{\'\i}as, Strelec, Thompson,
  Waites, Wiggins, and Zaragoza}]{banon-etal-2020-paracrawl}
Marta Ba{\~n}{\'o}n, Pinzhen Chen, Barry Haddow, Kenneth Heafield, Hieu Hoang,
  Miquel Espl{\`a}-Gomis, Mikel~L. Forcada, Amir Kamran, Faheem Kirefu, Philipp
  Koehn, Sergio Ortiz~Rojas, Leopoldo Pla~Sempere, Gema
  Ram{\'\i}rez-S{\'a}nchez, Elsa Sarr{\'\i}as, Marek Strelec, Brian Thompson,
  William Waites, Dion Wiggins, and Jaume Zaragoza. 2020.
\newblock \href {https://doi.org/10.18653/v1/2020.acl-main.417} {{P}ara{C}rawl:
  Web-scale acquisition of parallel corpora}.
\newblock In \emph{Proceedings of the 58th Annual Meeting of the Association
  for Computational Linguistics}, pages 4555--4567, Online. Association for
  Computational Linguistics.

\bibitem[{Bommasani et~al.(2021)Bommasani, Hudson, and
  et~al.}]{Bommasani2021OnTO}
Rishi Bommasani, Drew~A. Hudson, and Ehsan~Adeli et~al. 2021.
\newblock On the opportunities and risks of foundation models.
\newblock \emph{ArXiv}, abs/2108.07258.

\bibitem[{Botha et~al.(2017)Botha, Pitler, Ma, Bakalov, Salcianu, Weiss,
  McDonald, and Petrov}]{botha-etal-2017-natural}
Jan~A. Botha, Emily Pitler, Ji~Ma, Anton Bakalov, Alex Salcianu, David Weiss,
  Ryan McDonald, and Slav Petrov. 2017.
\newblock \href {https://doi.org/10.18653/v1/D17-1309} {Natural language
  processing with small feed-forward networks}.
\newblock In \emph{Proceedings of the 2017 Conference on Empirical Methods in
  Natural Language Processing}, pages 2879--2885, Copenhagen, Denmark.
  Association for Computational Linguistics.

\bibitem[{Broder(1997)}]{Broder1997On}
A.~Broder. 1997.
\newblock On the resemblance and containment of documents.
\newblock In \emph{Proceedings of the Compression and Complexity of Sequences}.

\bibitem[{Brown et~al.(2020)Brown, Mann, and et~al.}]{Brown2020LanguageMA}
Tom Brown, Benjamin Mann, and et~al. 2020.
\newblock Language models are few-shot learners.
\newblock \emph{ArXiv}, abs/2005.14165.

\bibitem[{Chowdhery et~al.(2022)Chowdhery, Narang, and
  et~al.}]{Chowdhery2022PaLMSL}
Aakanksha Chowdhery, Sharan Narang, and Jacob~Devlin et~al. 2022.
\newblock Palm: Scaling language modeling with pathways.
\newblock \emph{ArXiv}, abs/2204.02311.

\bibitem[{Computer(2023)}]{together2023redpajama}
Together Computer. 2023.
\newblock \href {https://github.com/togethercomputer/RedPajama-Data}
  {Redpajama: An open source recipe to reproduce llama training dataset}.

\bibitem[{Conneau et~al.(2020)Conneau, Khandelwal, Goyal, Chaudhary, Wenzek,
  Guzm{\'a}n, Grave, Ott, Zettlemoyer, and
  Stoyanov}]{conneau-etal-2020-unsupervised}
Alexis Conneau, Kartikay Khandelwal, Naman Goyal, Vishrav Chaudhary, Guillaume
  Wenzek, Francisco Guzm{\'a}n, Edouard Grave, Myle Ott, Luke Zettlemoyer, and
  Veselin Stoyanov. 2020.
\newblock \href {https://doi.org/10.18653/v1/2020.acl-main.747} {Unsupervised
  cross-lingual representation learning at scale}.
\newblock In \emph{Proceedings of the 58th Annual Meeting of the Association
  for Computational Linguistics}, pages 8440--8451, Online. Association for
  Computational Linguistics.

\bibitem[{Dekking et~al.(2007)Dekking, Kraaikamp, Paul, and
  Meester}]{Dekking2007AMI}
Michel Dekking, Cornelis Kraaikamp, Hendrik Paul, and Ludolf~Erwin Meester.
  2007.
\newblock A modern introduction to probability and statistics: Understanding
  why and how.
\newblock In \emph{Springer Texts in Statistics}.

\bibitem[{Devlin et~al.(2019)Devlin, Chang, Lee, and
  Toutanova}]{devlin-etal-2019-bert}
Jacob Devlin, Ming-Wei Chang, Kenton Lee, and Kristina Toutanova. 2019.
\newblock \href {https://doi.org/10.18653/v1/N19-1423} {{BERT}: Pre-training of
  deep bidirectional transformers for language understanding}.
\newblock In \emph{Proceedings of the 2019 Conference of the North {A}merican
  Chapter of the Association for Computational Linguistics: Human Language
  Technologies, Volume 1 (Long and Short Papers)}, pages 4171--4186,
  Minneapolis, Minnesota. Association for Computational Linguistics.

\bibitem[{Gao et~al.(2020)Gao, Biderman, Black, Golding, Hoppe, Foster, Phang,
  He, Thite, Nabeshima, Presser, and Leahy}]{Gao2020ThePA}
Leo Gao, Stella~Rose Biderman, Sid Black, Laurence Golding, Travis Hoppe,
  Charles Foster, Jason Phang, Horace He, Anish Thite, Noa Nabeshima, Shawn
  Presser, and Connor Leahy. 2020.
\newblock The pile: An 800gb dataset of diverse text for language modeling.
\newblock \emph{ArXiv}, abs/2101.00027.

\bibitem[{Grave et~al.(2018)Grave, Bojanowski, Gupta, Joulin, and
  Mikolov}]{grave-etal-2018-learning}
Edouard Grave, Piotr Bojanowski, Prakhar Gupta, Armand Joulin, and Tomas
  Mikolov. 2018.
\newblock \href {https://aclanthology.org/L18-1550} {Learning word vectors for
  157 languages}.
\newblock In \emph{Proceedings of the Eleventh International Conference on
  Language Resources and Evaluation ({LREC} 2018)}, Miyazaki, Japan. European
  Language Resources Association (ELRA).

\bibitem[{Heafield(2011)}]{heafield-2011-kenlm}
Kenneth Heafield. 2011.
\newblock \href {https://aclanthology.org/W11-2123} {{K}en{LM}: Faster and
  smaller language model queries}.
\newblock In \emph{Proceedings of the Sixth Workshop on Statistical Machine
  Translation}, pages 187--197, Edinburgh, Scotland. Association for
  Computational Linguistics.

\bibitem[{Hernandez et~al.(2022)Hernandez, Brown, Conerly, DasSarma, Drain,
  El-Showk, Elhage, Hatfield-Dodds, Henighan, Hume, Johnston, Mann, Olah,
  Olsson, Amodei, Joseph, Kaplan, and McCandlish}]{Hernandez2022ScalingLA}
Danny Hernandez, Tom~B. Brown, Tom Conerly, Nova DasSarma, Dawn Drain, Sheer
  El-Showk, Nelson Elhage, Zac Hatfield-Dodds, T.~J. Henighan, Tristan Hume,
  Scott Johnston, Benjamin Mann, Christopher Olah, Catherine Olsson, Dario
  Amodei, Nicholas Joseph, Jared Kaplan, and Sam McCandlish. 2022.
\newblock \href {https://api.semanticscholar.org/CorpusID:248986979} {Scaling
  laws and interpretability of learning from repeated data}.
\newblock \emph{ArXiv}, abs/2205.10487.

\bibitem[{Holtzman et~al.(2019)Holtzman, Buys, Du, Forbes, and
  Choi}]{Holtzman2019TheCC}
Ari Holtzman, Jan Buys, Li~Du, Maxwell Forbes, and Yejin Choi. 2019.
\newblock The curious case of neural text degeneration.
\newblock \emph{ArXiv}, abs/1904.09751.

\bibitem[{Joulin et~al.(2016)Joulin, Grave, Bojanowski, Douze, J{\'e}gou, and
  Mikolov}]{Joulin2016FastTextzipCT}
Armand Joulin, Edouard Grave, Piotr Bojanowski, Matthijs Douze, Herv{\'e}
  J{\'e}gou, and Tomas Mikolov. 2016.
\newblock Fasttext.zip: Compressing text classification models.
\newblock \emph{ArXiv}, abs/1612.03651.

\bibitem[{Kenton et~al.(2021)Kenton, Everitt, Weidinger, Gabriel, Mikulik, and
  Irving}]{Kenton2021AlignmentOL}
Zachary Kenton, Tom Everitt, Laura Weidinger, Iason Gabriel, Vladimir Mikulik,
  and Geoffrey Irving. 2021.
\newblock Alignment of language agents.
\newblock \emph{ArXiv}, abs/2103.14659.

\bibitem[{Kreutzer et~al.(2022)Kreutzer, Caswell, Wang, Wahab, van Esch,
  Ulzii-Orshikh, Tapo, Subramani, Sokolov, Sikasote, Setyawan, Sarin, Samb,
  Sagot, Rivera, Rios, Papadimitriou, Osei, Suarez, Orife, Ogueji, Rubungo,
  Nguyen, M{\"u}ller, M{\"u}ller, Muhammad, Muhammad, Mnyakeni, Mirzakhalov,
  Matangira, Leong, Lawson, Kudugunta, Jernite, Jenny, Firat, Dossou, Dlamini,
  de~Silva, {\c{C}}abuk~Ball{\i}, Biderman, Battisti, Baruwa, Bapna, Baljekar,
  Azime, Awokoya, Ataman, Ahia, Ahia, Agrawal, and
  Adeyemi}]{kreutzer-etal-2022-quality}
Julia Kreutzer, Isaac Caswell, Lisa Wang, Ahsan Wahab, Daan van Esch,
  Nasanbayar Ulzii-Orshikh, Allahsera Tapo, Nishant Subramani, Artem Sokolov,
  Claytone Sikasote, Monang Setyawan, Supheakmungkol Sarin, Sokhar Samb,
  Beno{\^\i}t Sagot, Clara Rivera, Annette Rios, Isabel Papadimitriou, Salomey
  Osei, Pedro~Ortiz Suarez, Iroro Orife, Kelechi Ogueji, Andre~Niyongabo
  Rubungo, Toan~Q. Nguyen, Mathias M{\"u}ller, Andr{\'e} M{\"u}ller,
  Shamsuddeen~Hassan Muhammad, Nanda Muhammad, Ayanda Mnyakeni, Jamshidbek
  Mirzakhalov, Tapiwanashe Matangira, Colin Leong, Nze Lawson, Sneha Kudugunta,
  Yacine Jernite, Mathias Jenny, Orhan Firat, Bonaventure F.~P. Dossou, Sakhile
  Dlamini, Nisansa de~Silva, Sakine {\c{C}}abuk~Ball{\i}, Stella Biderman,
  Alessia Battisti, Ahmed Baruwa, Ankur Bapna, Pallavi Baljekar, Israel~Abebe
  Azime, Ayodele Awokoya, Duygu Ataman, Orevaoghene Ahia, Oghenefego Ahia,
  Sweta Agrawal, and Mofetoluwa Adeyemi. 2022.
\newblock \href {https://doi.org/10.1162/tacl_a_00447} {Quality at a glance: An
  audit of web-crawled multilingual datasets}.
\newblock \emph{Transactions of the Association for Computational Linguistics},
  10:50--72.

\bibitem[{Kudo(2018)}]{kudo-2018-subword}
Taku Kudo. 2018.
\newblock \href {https://doi.org/10.18653/v1/P18-1007} {Subword regularization:
  Improving neural network translation models with multiple subword
  candidates}.
\newblock In \emph{Proceedings of the 56th Annual Meeting of the Association
  for Computational Linguistics (Volume 1: Long Papers)}, pages 66--75,
  Melbourne, Australia. Association for Computational Linguistics.

\bibitem[{Lai et~al.(2023)Lai, Ngo, Veyseh, Man, Dernoncourt, Bui, and
  Nguyen}]{Lai2023ChatGPTBE}
Viet~Dac Lai, Nghia~Trung Ngo, Amir Pouran~Ben Veyseh, Hieu Man, Franck
  Dernoncourt, Trung Bui, and Thien~Huu Nguyen. 2023.
\newblock Chatgpt beyond english: Towards a comprehensive evaluation of large
  language models in multilingual learning.
\newblock \emph{ArXiv}, abs/2304.05613.

\bibitem[{Lauren{\c{c}}on et~al.(2022)Lauren{\c{c}}on, Saulnier, Wang, Akiki,
  del Moral, Scao, Werra, Mou, Ponferrada, Nguyen, Frohberg, {\v{S}}a{\v{s}}ko,
  Lhoest, McMillan-Major, Dupont, Biderman, Rogers, allal, Toni, Pistilli,
  Nguyen, Nikpoor, Masoud, Colombo, de~la Rosa, Villegas, Thrush, Longpre,
  Nagel, Weber, Mu{\~n}oz, Zhu, Strien, Alyafeai, Almubarak, Chien,
  Gonzalez-Dios, Soroa, Lo, Dey, Suarez, Gokaslan, Bose, Adelani, Phan, Tran,
  Yu, Pai, Chim, Lepercq, Ilic, Mitchell, Luccioni, and
  Jernite}]{laurenon2022the}
Hugo Lauren{\c{c}}on, Lucile Saulnier, Thomas Wang, Christopher Akiki,
  Albert~Villanova del Moral, Teven~Le Scao, Leandro~Von Werra, Chenghao Mou,
  Eduardo~Gonz{\'a}lez Ponferrada, Huu Nguyen, J{\"o}rg Frohberg, Mario
  {\v{S}}a{\v{s}}ko, Quentin Lhoest, Angelina McMillan-Major, G{\'e}rard
  Dupont, Stella Biderman, Anna Rogers, Loubna~Ben allal, Francesco~De Toni,
  Giada Pistilli, Olivier Nguyen, Somaieh Nikpoor, Maraim Masoud, Pierre
  Colombo, Javier de~la Rosa, Paulo Villegas, Tristan Thrush, Shayne Longpre,
  Sebastian Nagel, Leon Weber, Manuel~Romero Mu{\~n}oz, Jian Zhu, Daniel~Van
  Strien, Zaid Alyafeai, Khalid Almubarak, Vu~Minh Chien, Itziar Gonzalez-Dios,
  Aitor Soroa, Kyle Lo, Manan Dey, Pedro~Ortiz Suarez, Aaron Gokaslan, Shamik
  Bose, David~Ifeoluwa Adelani, Long Phan, Hieu Tran, Ian Yu, Suhas Pai, Jenny
  Chim, Violette Lepercq, Suzana Ilic, Margaret Mitchell, Sasha Luccioni, and
  Yacine Jernite. 2022.
\newblock The bigscience {ROOTS} corpus: A 1.6{TB} composite multilingual
  dataset.
\newblock In \emph{Thirty-sixth Conference on Neural Information Processing
  Systems Datasets and Benchmarks Track}.

\bibitem[{Lee et~al.(2022)Lee, Ippolito, Nystrom, Zhang, Eck, Callison-Burch,
  and Carlini}]{lee-etal-2022-deduplicating}
Katherine Lee, Daphne Ippolito, Andrew Nystrom, Chiyuan Zhang, Douglas Eck,
  Chris Callison-Burch, and Nicholas Carlini. 2022.
\newblock \href {https://doi.org/10.18653/v1/2022.acl-long.577} {Deduplicating
  training data makes language models better}.
\newblock In \emph{Proceedings of the 60th Annual Meeting of the Association
  for Computational Linguistics (Volume 1: Long Papers)}, pages 8424--8445,
  Dublin, Ireland. Association for Computational Linguistics.

\bibitem[{Leskovec et~al.(2020)Leskovec, Rajaraman, and Ullman}]{Leskovec2020}
Jure Leskovec, Anand Rajaraman, and Jeffrey~David Ullman. 2020.
\newblock \href {http://infolab.stanford.edu/~ullman/mmds/ch3.pdf} {Mining of
  massive datasets}.
\newblock In \emph{Cambridge University Press}.

\bibitem[{Lewis et~al.(2020)Lewis, Liu, Goyal, Ghazvininejad, Mohamed, Levy,
  Stoyanov, and Zettlemoyer}]{lewis-etal-2020-bart}
Mike Lewis, Yinhan Liu, Naman Goyal, Marjan Ghazvininejad, Abdelrahman Mohamed,
  Omer Levy, Veselin Stoyanov, and Luke Zettlemoyer. 2020.
\newblock \href {https://doi.org/10.18653/v1/2020.acl-main.703} {{BART}:
  Denoising sequence-to-sequence pre-training for natural language generation,
  translation, and comprehension}.
\newblock In \emph{Proceedings of the 58th Annual Meeting of the Association
  for Computational Linguistics}, pages 7871--7880, Online. Association for
  Computational Linguistics.

\bibitem[{Lieber et~al.(2021)Lieber, Sharir, Lenz, and
  Shoham}]{Lieber2021Jurassic}
Opher Lieber, Or~Sharir, Barak Lenz, and Yoav Shoham. 2021.
\newblock Jurassic-1: Technical details and evaluation.
\newblock \emph{White Paper. AI21 Labs.}

\bibitem[{Liu et~al.(2019)Liu, Ott, Goyal, Du, Joshi, Chen, Levy, Lewis,
  Zettlemoyer, and Stoyanov}]{Liu2019RoBERTaAR}
Yinhan Liu, Myle Ott, Naman Goyal, Jingfei Du, Mandar Joshi, Danqi Chen, Omer
  Levy, Mike Lewis, Luke Zettlemoyer, and Veselin Stoyanov. 2019.
\newblock Roberta: A robustly optimized bert pretraining approach.
\newblock \emph{ArXiv}, abs/1907.11692.

\bibitem[{MosaicML(2023)}]{MTP}
MosaicML. 2023.
\newblock Introducing mpt-7b: A new standard for open-source, commercially
  usable llms.
\newblock \emph{\url{https://www.mosaicml.com/blog/mpt-7b}}.

\bibitem[{Nagel()}]{ccnews}
Sebastian Nagel.
\newblock Cc-news. http: //web.archive.org/save/http:
  //commoncrawl.org/2016/10/news- dataset-available.

\bibitem[{Ortiz~Su{\'a}rez et~al.(2020)Ortiz~Su{\'a}rez, Romary, and
  Sagot}]{ortiz-suarez-etal-2020-monolingual}
Pedro~Javier Ortiz~Su{\'a}rez, Laurent Romary, and Beno{\^\i}t Sagot. 2020.
\newblock \href {https://doi.org/10.18653/v1/2020.acl-main.156} {A monolingual
  approach to contextualized word embeddings for mid-resource languages}.
\newblock In \emph{Proceedings of the 58th Annual Meeting of the Association
  for Computational Linguistics}, pages 1703--1714, Online. Association for
  Computational Linguistics.

\bibitem[{{Ortiz Su{\'a}rez} et~al.(2019){Ortiz Su{\'a}rez}, Sagot, and
  Romary}]{OrtizSuarezSagotRomary2019}
Pedro~Javier {Ortiz Su{\'a}rez}, Beno{\^i}t Sagot, and Laurent Romary. 2019.
\newblock \href {http://nbn-resolving.de/urn:nbn:de:bsz:mh39-90215}
  {Asynchronous pipelines for processing huge corpora on medium to low resource
  infrastructures}.
\newblock In \emph{Proceedings of the Workshop on Challenges in the Management
  of Large Corpora (CMLC-7) 2019. Cardiff, 22nd July 2019}.

\bibitem[{Penedo et~al.(2023)Penedo, Malartic, Hesslow, Cojocaru, Cappelli,
  Alobeidli, Pannier, Almazrouei, and Launay}]{Penedo2023TheRD}
Guilherme Penedo, Quentin Malartic, Daniel Hesslow, Ruxandra-Aim{\'e}e
  Cojocaru, Alessandro Cappelli, Hamza Alobeidli, Baptiste Pannier, Ebtesam
  Almazrouei, and Julien Launay. 2023.
\newblock \href {https://api.semanticscholar.org/CorpusID:259063761} {The
  refinedweb dataset for falcon llm: Outperforming curated corpora with web
  data, and web data only}.
\newblock \emph{ArXiv}, abs/2306.01116.

\bibitem[{Radford et~al.(2019)Radford, Wu, Child, Luan, Amodei, Sutskever
  et~al.}]{Radford2019Language}
Alec Radford, Jeffrey Wu, Rewon Child, David Luan, Dario Amodei, Ilya
  Sutskever, et~al. 2019.
\newblock Language models are unsupervised multitask learners.
\newblock \emph{OpenAI blog}.

\bibitem[{Rae et~al.(2021)Rae, Borgeaud, and et~al.}]{Rae2021ScalingLM}
Jack Rae, Sebastian Borgeaud, and et~al. 2021.
\newblock Scaling language models: Methods, analysis \& insights from training
  gopher.
\newblock \emph{ArXiv}, abs/2112.11446.

\bibitem[{Raffel et~al.(2020)Raffel, Shazeer, Roberts, Lee, Narang, Matena,
  Zhou, Li, and Liu}]{Raffel2020Xxploring}
Colin Raffel, Noam Shazeer, Adam Roberts, Katherine Lee, Sharan Narang, Michael
  Matena, Yanqi Zhou, Wei Li, and Peter~J. Liu. 2020.
\newblock Exploring the limits of transfer learning with a unified text-to-text
  transformer.
\newblock In \emph{Journal of Machine Learning Research}.

\bibitem[{Scao et~al.(2022)Scao, Fan, and et~al.}]{Scao2022BLOOMA1}
Teven Scao, Angela Fan, and et~al. 2022.
\newblock Bloom: A 176b-parameter open-access multilingual language model.
\newblock \emph{ArXiv}, abs/2211.05100.

\bibitem[{Schwenk et~al.(2021)Schwenk, Chaudhary, Sun, Gong, and
  Guzm{\'a}n}]{schwenk-etal-2021-wikimatrix}
Holger Schwenk, Vishrav Chaudhary, Shuo Sun, Hongyu Gong, and Francisco
  Guzm{\'a}n. 2021.
\newblock \href {https://doi.org/10.18653/v1/2021.eacl-main.115}
  {{W}iki{M}atrix: Mining 135{M} parallel sentences in 1620 language pairs from
  {W}ikipedia}.
\newblock In \emph{Proceedings of the 16th Conference of the European Chapter
  of the Association for Computational Linguistics: Main Volume}, pages
  1351--1361, Online. Association for Computational Linguistics.

\bibitem[{Shoeybi et~al.(2019)Shoeybi, Patwary, Puri, LeGresley, Casper, and
  Catanzaro}]{Shoeybi2019MegatronLMTM}
Mohammad Shoeybi, Mostofa Patwary, Raul Puri, Patrick LeGresley, Jared Casper,
  and Bryan Catanzaro. 2019.
\newblock Megatron-lm: Training multi-billion parameter language models using
  model parallelism.
\newblock \emph{ArXiv}, abs/1909.08053.

\bibitem[{Tamkin et~al.(2021)Tamkin, Brundage, Clark, and
  Ganguli}]{Tamkin2021UnderstandingTC}
Alex Tamkin, Miles Brundage, Jack Clark, and Deep Ganguli. 2021.
\newblock Understanding the capabilities, limitations, and societal impact of
  large language models.
\newblock \emph{ArXiv}, abs/2102.02503.

\bibitem[{Touvron et~al.(2023)Touvron, Lavril, and et~al.}]{Touvron2023LLaMAOA}
Hugo Touvron, Thibaut Lavril, and Gautier~Izacard et~al. 2023.
\newblock Llama: Open and efficient foundation language models.
\newblock \emph{ArXiv}, abs/2302.13971.

\bibitem[{Trinh and Le(2018)}]{Trinh2018ASM}
Trieu~H. Trinh and Quoc~V. Le. 2018.
\newblock A simple method for commonsense reasoning.
\newblock \emph{ArXiv}, abs/1806.02847.

\bibitem[{Vaswani et~al.(2017)Vaswani, Shazeer, Parmar, Uszkoreit, Jones,
  Gomez, Kaiser, and Polosukhin}]{Vaswani2017Attention}
Ashish Vaswani, Noam Shazeer, Niki Parmar, Jakob Uszkoreit, Llion Jones,
  Aidan~N Gomez, Lukasz Kaiser, and Illia Polosukhin. 2017.
\newblock Attention is all you need.
\newblock In \emph{Advances in Neural Information Processing Systems}.

\bibitem[{Wei et~al.(2022)Wei, Tay, Bommasani, Raffel, Zoph, Borgeaud,
  Yogatama, Bosma, Zhou, Metzler, hsin Chi, Hashimoto, Vinyals, Liang, Dean,
  and Fedus}]{Wei2022EmergentAO}
Jason Wei, Yi~Tay, Rishi Bommasani, Colin Raffel, Barret Zoph, Sebastian
  Borgeaud, Dani Yogatama, Maarten Bosma, Denny Zhou, Donald Metzler, Ed~Huai
  hsin Chi, Tatsunori Hashimoto, Oriol Vinyals, Percy Liang, Jeff Dean, and
  William Fedus. 2022.
\newblock Emergent abilities of large language models.
\newblock \emph{Transactions on Machine Learning Research}.

\bibitem[{Wei et~al.(2023)Wei, Wei, Lin, Li, Zhang, Ren, Li, Wan, Cao, Xie, Hu,
  Li, Hui, Yu, Liu, Yang, Huang, and Xie}]{Wei2023PolyLMAO}
Xiangpeng Wei, Hao-Ran Wei, Huan Lin, Tianhao Li, Pei Zhang, Xingzhang Ren, Mei
  Li, Yu~Wan, Zhiwei Cao, Binbin Xie, Tianxiang Hu, Shangjie Li, Binyuan Hui,
  Bowen Yu, Dayiheng Liu, Baosong Yang, Fei Huang, and Jun Xie. 2023.
\newblock Polylm: An open source polyglot large language model.
\newblock \emph{ArXiv}, abs/2307.06018.

\bibitem[{Weidinger et~al.(2021)Weidinger, Mellor, and
  et~al.}]{Weidinger2021EthicalAS}
Laura Weidinger, John F.~J. Mellor, and Maribeth~Rauh et~al. 2021.
\newblock Ethical and social risks of harm from language models.
\newblock \emph{ArXiv}, abs/2112.04359.

\bibitem[{Wenzek et~al.(2020)Wenzek, Lachaux, Conneau, Chaudhary, Guzm{\'a}n,
  Joulin, and Grave}]{wenzek-etal-2020-ccnet}
Guillaume Wenzek, Marie-Anne Lachaux, Alexis Conneau, Vishrav Chaudhary,
  Francisco Guzm{\'a}n, Armand Joulin, and Edouard Grave. 2020.
\newblock \href {https://aclanthology.org/2020.lrec-1.494} {{CCN}et: Extracting
  high quality monolingual datasets from web crawl data}.
\newblock In \emph{Proceedings of the Twelfth Language Resources and Evaluation
  Conference}, pages 4003--4012, Marseille, France. European Language Resources
  Association.

\bibitem[{Xue et~al.(2021)Xue, Constant, Roberts, Kale, Al-Rfou, Siddhant,
  Barua, and Raffel}]{xue-etal-2021-mt5}
Linting Xue, Noah Constant, Adam Roberts, Mihir Kale, Rami Al-Rfou, Aditya
  Siddhant, Aditya Barua, and Colin Raffel. 2021.
\newblock \href {https://doi.org/10.18653/v1/2021.naacl-main.41} {m{T}5: A
  massively multilingual pre-trained text-to-text transformer}.
\newblock In \emph{Proceedings of the 2021 Conference of the North American
  Chapter of the Association for Computational Linguistics: Human Language
  Technologies}, pages 483--498, Online. Association for Computational
  Linguistics.

\bibitem[{Zhu et~al.(2015)Zhu, Kiros, Zemel, Salakhutdinov, Urtasun, Torralba,
  and Fidler}]{Zhu2015AligningBA}
Yukun Zhu, Ryan Kiros, Richard~S. Zemel, Ruslan Salakhutdinov, Raquel Urtasun,
  Antonio Torralba, and Sanja Fidler. 2015.
\newblock Aligning books and movies: Towards story-like visual explanations by
  watching movies and reading books.
\newblock \emph{Proceedings of the IEEE International Conference on Computer
  Vision (ICCV)}.

\end{thebibliography}
\bibliographystyle{acl_natbib}

\clearpage

\appendix

\end{document}